\documentclass{article} 

\usepackage{iclr2020_conference,times}
\usepackage{float}
\usepackage{amsmath,amsfonts,bm}

\def\eqref#1{equation~\ref{#1}}
\def\1{\bm{1}}

\DeclareMathAlphabet{\mathsfit}{\encodingdefault}{\sfdefault}{m}{sl}
\SetMathAlphabet{\mathsfit}{bold}{\encodingdefault}{\sfdefault}{bx}{n}

\usepackage{bbm}
\usepackage{hyperref}
\usepackage{color}
\usepackage{url}
\usepackage{multirow}
\usepackage{graphicx} %use graph format
\usepackage{epstopdf}
\usepackage{array}
\usepackage{amsthm,amsmath,amssymb}
\usepackage{mathrsfs}
\usepackage{algorithm}
\usepackage{algorithmic}
\usepackage{bm}
\makeatletter
\newcommand{\figcaption}{\def\@captype{figure}\caption}
\newcommand{\tabcaption}{\def\@captype{table}\caption}
\makeatother

\title{Mutual Information Gradient Estimation for  Representation Learning}

\author{Liangjian Wen$^{1,2}$, Yiji Zhou$^1$, Lirong He$^1$, Mingyuan Zhou$^3$, Zenglin Xu$^{4,2,1}$\\
$^1$ SMILE Lab, School of Computer Science and Engineering \\
University of Electronic Science and Technology of China, Chengdu, China \\
$^2$ Center for Artificial Intelligence \\
Peng Cheng Laboratory, 
Shenzhen, China \\ 
$^3$ McCombs School of Business \\
University of Texas at Austin, 
Austin, United States \\ 
$^4$ School of Computer Science and Technology \\
Harbin Institute of Technology,
Shenzhen, China \\ 
\texttt{wlj6816@gmail.com,zhouyiji@outlook.com,ronghe1217@gmail.com,} \\
\texttt{mingyuan.zhou@mccombs.utexas.edu,xuzenglin@hit.edu.cn} \\
}
\iclrfinalcopy % Uncomment for camera-ready version, but NOT for submission.
\begin{document}

\maketitle

\begin{abstract}
Mutual Information (MI) plays an important role in representation learning. However, MI is unfortunately intractable in continuous and high-dimensional settings. Recent advances establish tractable and scalable MI estimators to discover useful representation. However, most of the existing methods are not capable of providing an accurate estimation of MI with low-variance when the MI is large. We argue that directly estimating the gradients of MI is more appealing for representation learning than estimating MI in itself. To this end, we propose the Mutual Information Gradient Estimator (MIGE) for representation learning based on the score estimation of implicit distributions. MIGE exhibits a tight and smooth gradient estimation of MI in the high-dimensional and large-MI settings. We expand the applications of MIGE in both unsupervised learning of deep representations based on InfoMax and the Information Bottleneck method. Experimental results have indicated significant performance improvement in learning useful representation.

\end{abstract}

\section{Introduction}
Mutual information (MI) is an appealing metric widely used in information theory and machine learning to quantify the amount of shared information between a pair of random variables. Specifically, given a pair of random variables $\mathbf{x},\mathbf{y}$, the MI, denoted by $I(\mathbf{x};\mathbf{y})$, is defined as
\begin{align}
    I(\mathbf{x};\mathbf{y})=\mathbb{E}_{p(\mathbf{x}, \mathbf{y})}\left[\log \frac{p(\mathbf{x} , \mathbf{y})}{p(\mathbf{x})p(\mathbf{y})}\right],
\end{align}
where $\mathbb{E}$ is the expectation over the given distribution.
Since MI is invariant to invertible and smooth transformations, it can capture non-linear statistical dependencies between variables~\citep{kinney2014equitability}.
These appealing properties make it act as a fundamental measure of true dependence.
Therefore, MI has found applications in a wide range of machine learning tasks, including feature selection~\citep{kwak2002input,fleuret2004fast,peng2005feature},  clustering~\citep{muller2012information,ver2015maximally}, and causality~\citep{butte1999mutual}. It has also been pervasively used in science, such as biomedical sciences~\citep{maes1997multimodality}, computational biology~\citep{krishnaswamy2014conditional}, and computational neuroscience~\citep{palmer2015predictive}.

Recently, there has been a revival of methods in unsupervised representation learning based on MI. A seminal work is the InfoMax principle \citep{linsker1988self}, where given an input instance $x$, the goal of the InfoMax principle is to learn a representation $E_{\psi}(x)$ by maximizing the MI between the input and its representation. 
A growing set of recent works have demonstrated promising empirical performance in unsupervised representation learning via MI maximization \citep{krause2010discriminative,hu2017learning,DBLP:conf/icml/AlemiPFDS018,oord2018representation,DBLP:conf/iclr/HjelmFLGBTB19}. Another closely related work is the Information Bottleneck method \citep{tishby2000information,DBLP:conf/iclr/AlemiFD017}, where
MI is used to limit the contents of representations. Specifically, the representations are learned by extracting task-related information from the original data while being constrained to discard parts that are irrelevant to the task. Several recent works have also suggested that by controlling the amount of information between learned representations and the original data, one can tune desired characteristics of trained models such as generalization error~\citep{tishby2015deep,vera2018role}, robustness~\citep{ DBLP:conf/iclr/AlemiFD017}, and detection of out-of-distribution data~\citep{alemi2018uncertainty}.
 
Despite playing a pivotal role across a variety of domains, MI is notoriously intractable.
Exact computation is only tractable for discrete variables, or for a limited family of problems where the probability distributions are known.
For more general problems, MI is challenging to analytically compute or estimate from samples.
A variety of MI estimators have been developed over the years, including likelihood-ratio estimators~\citep{suzuki2008approximating}, binning~\citep{fraser1986independent,darbellay1999estimation,shwartz2017opening}, k-nearest neighbors~\citep{kozachenko1987sample,kraskov2004estimating,perez2008kullback,singh2016finite}, and kernel density estimators~\citep{moon1995estimation,kwak2002input,kandasamy2015nonparametric}.
However, few of these mutual information estimators scale well with dimension and sample size in machine learning problems~\citep{gao2015efficient}.

In order to overcome the intractability of MI in the continuous and high-dimensional settings, \cite{DBLP:conf/iclr/AlemiFD017} combines variational bounds of \cite{barber2003algorithm} with neural networks for the estimation. However, the tractable density for the approximate distribution is required due to variational approximation. This limits its application to the general-purpose estimation, since the underlying distributions are often unknown. Alternatively, the Mutual Information Neural Estimation (MINE,~\cite{belghazi2018mine}) and the Jensen-Shannon MI estimator (JSD,~\cite{DBLP:conf/iclr/HjelmFLGBTB19}) enable differentiable and tractable estimation of MI by training a discriminator to distinguish samples coming from the joint distribution or the product of the marginals.
In detail, MINE employs a lower-bound to the MI based on the Donsker-Varadhan representation of the KL-divergence, and JSD follows the formulation of f-GAN KL-divergence.
In general, these estimators are often noisy and can lead to unstable training due to their dependence on the discriminator used to estimate the bounds of mutual information. As pointed out by \cite{DBLP:conf/icml/PooleOOAT19}, these unnormalized critic estimators of MI exhibit high variance and are challenging to tune for estimation. An alternative low-variance choice of MI estimator is Information Noise-Contrastive Estimation (InfoNCE,~\cite{oord2018representation}), 
which introduces the Noise-Contrastive Estimation with flexible critics parameterized by neural networks as a bound to approximate MI.
Nonetheless, its estimation saturates at log of the batch size and suffers from high bias. Despite their modeling power, none of the estimators are capable of providing accurate estimation of MI with low variance when the MI is large and the batch size is small~\citep{DBLP:conf/icml/PooleOOAT19}.  
As supported by the theoretical findings in ~\cite{mcallester2018formal},
any distribution-free high-confidence lower bound on entropy requires a sample size exponential in the size of the bound. More discussions about the bounds of MI and their relationship can be referred to \cite{DBLP:conf/icml/PooleOOAT19}. 

% As indicated in \cite{Tschannen2019OnMI},  maximizing tighter bounds on MI can result in worse representations. Thus, we propose to estimate the gradients of mutual information instead of the MI. 

In summary,
existing estimators first approximate MI and then use these approximations to optimize the associated parameters. 
%During the practical optimization, we care about computing the gradients of MI.
For estimating MI based on any finite number of samples, there exists an infinite number of functions, with arbitrarily diverse gradients, that can perfectly approximate the true MI at these samples. 
However, these approximate functions can lead to unstable training and poor performance in optimization due to gradients discrepancy between approximate estimation and true MI. Estimating gradients of MI rather than estimating MI may be a better approach for MI optimization. 
To this end, to the best of our knowledge, we firstly propose the Mutual Information Gradient Estimator (MIGE) in representation learning.
In detail, we estimate the score function of an implicit distribution, $\nabla_{\mathbf{x}} \log q(\mathbf{x})$, to achieve a general-purpose MI gradient estimation for representation learning. In particular, to deal with high-dimensional inputs, such as text, images and videos, score function estimation via Spectral Stein Gradient Estimator (SSGE) ~\citep{shi2018spectral} is computationally expensive and complex. We thus propose an efficient high-dimensional score function estimator to make SSGE scalable. To this end, we derive a new reparameterization trick for the representation distribution based on the lower-variance reparameterization trick proposed by~\cite{roeder2017sticking}.  

\begin{figure}
    \centering
    \includegraphics[width=5.5in, keepaspectratio]{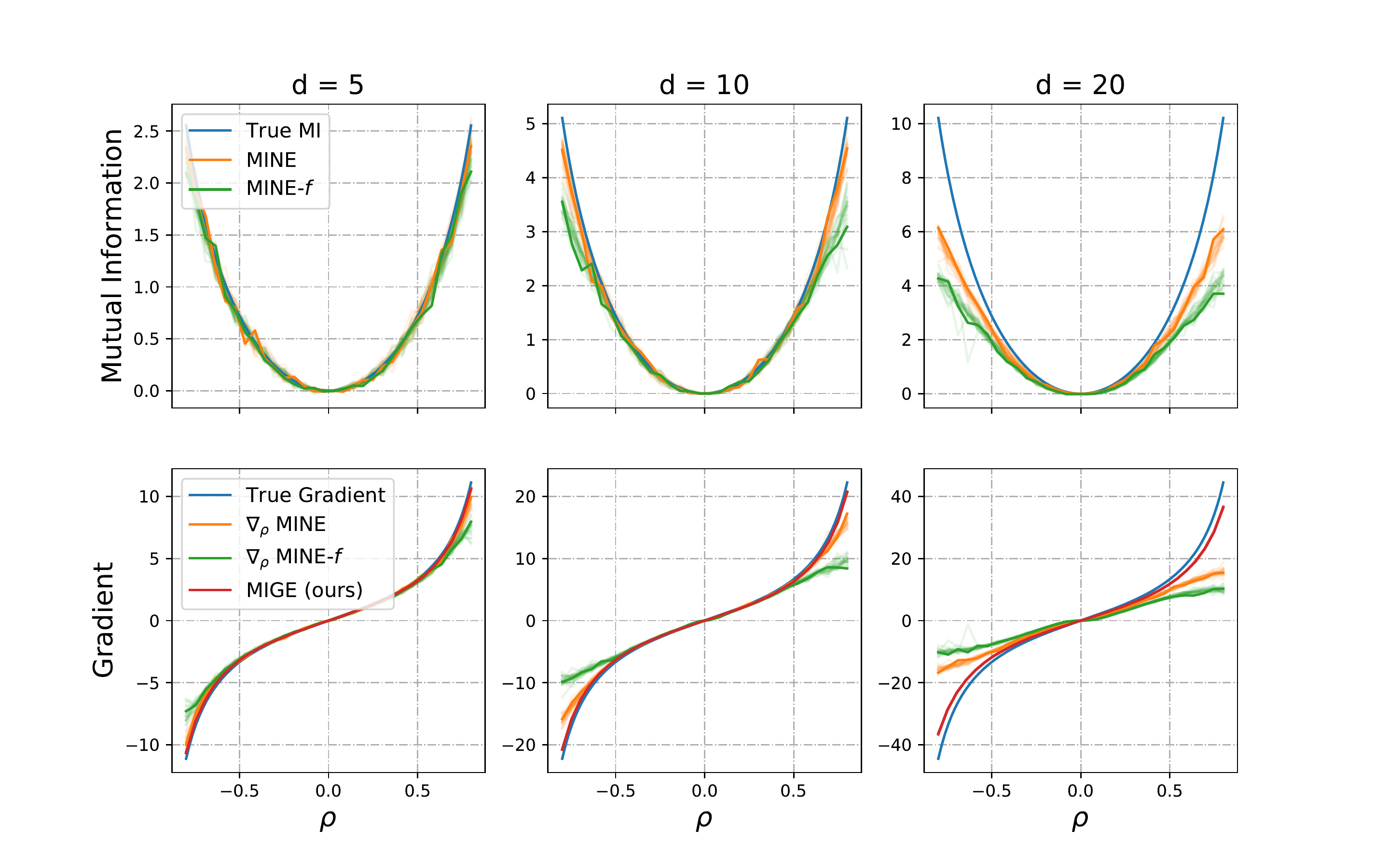}
    \caption{Estimation performance of MINE, MINE-$f$ and MIGE. Each estimation approach has been taken additional 20 times and plotted with light curves. \textbf{Top:} True MI and corresponding estimation of MINE and MINE-$f$. \textbf{Bottom:} True gradient and corresponding estimation of MINE, MINE-$f$ and MIGE. Our approach MIGE only appears in bottom figures since it directly gives gradient estimation. As we observe, MIGE gives more stable, smooth and accurate results.}
    \label{fig:toy}
\end{figure}

We summarize the contributions of this paper as follows:
\begin{itemize}
	\item We propose the Mutual Information Gradient Estimator (MIGE) for representation learning based on the score function estimation of implicit distributions. Compared with MINE and MINE-$f$, MIGE provides a tighter and smoother gradient estimation of MI in a high-dimensional and large-MI setting, as shown in Figure~\ref{fig:toy} of Section \ref{sec:toy}.
	\item We propose the Scalable SSGE to alleviate the exorbitant computational cost of SSGE in high-dimensional settings.
	\item To learn meaningful representations, we apply SSGE as gradient estimators for both InfoMax and Information Bottlenck, and have achieved improved performance than their corresponding competitors.
	%\item We present a gradient estimation solution to the unsupervised representation learning based on InfoMax, which significantly improves the performance of deep information models.
%	\item We present a gradient estimator of the Information Bottleneck (IB) method with MIGE in a continuous setting. Experimental results have indicated that our method outperforms variational IB methods and MINE IB methods.
\end{itemize}

\section{Scalable Spectral Stein Gradient Estimator}
Score estimation of implicit distributions has been widely explored in the past few years \citep{song2019sliced,li2017gradient,shi2018spectral}. 
A promising method of score estimation is the Stein gradient estimator \citep{li2017gradient,shi2018spectral}, which is proposed for implicit distributions. It is inspired by generalized Stein’s identity ~\citep{gorham2015measuring,liu2016stein} as follows.

\textbf{Stein’s identity.} Let $q(\mathbf{x})$ be a continuously differentiable (also called smooth) density supported on $\mathcal{X} \subseteq \mathbb{R}^{d}$, and $\boldsymbol{h}(\mathbf{x})=\left[h_{1}(\mathbf{x}), h_{2}(\mathbf{x}), \ldots, h_{d^{\prime}}(\mathbf{x})\right]^{\mathrm{T}}$ is a smooth vector function. Further, the boundary conditions on $\boldsymbol{h}$ is:
\begin{align}
q(\mathbf{x})\boldsymbol{h}(\mathbf{x})=0, \forall \mathbf{x} \in \partial \mathcal{X} \text { if } \mathcal{X} \text { is compact}, \text {or}~~~ \lim _{\mathbf{x} \rightarrow \infty} q(\mathbf{x}) \boldsymbol{h}(\mathbf{x})=0 \text { if } \mathcal{X}=\mathbb{R}^{d}.
\end{align}
Under this condition, the following identity can be easily checked using integration by parts, assuming mild zero boundary conditions on $\boldsymbol{h}$,
\begin{align}\label{steinid}
\mathbb{E}_{q}\left[\boldsymbol{h}(\mathbf{x}) \nabla_{\mathbf{x}} \log q(\mathbf{x})^{\mathrm{T}}+\nabla_{\mathbf{x}} \boldsymbol{h}(\mathbf{x})\right]=\mathbf{0}.
\end{align}
Here $\boldsymbol{h}$ is called the Stein class of $q(\boldsymbol{x})$  if Stein’s identity Eq.~(\ref{steinid})  holds.
Monte Carlo estimation of the expectation in Eq.~(\ref{steinid}) builds the connection between $\nabla_{\mathbf{x}} \log q(\mathbf{x})$ and the samples from $q(\mathbf{x})$ in Stein’s identity. 
For modeling implicit distributions, 
Motivated by Stein’s identity, \cite{shi2018spectral} proposed Spectral Stein Gradient Estimator (SSGE) for implicit distributions based on Stein's identity and a spectral decomposition of kernel operators where the eigenfunctions are approximated by the Nystr$\ddot{o}$m method. Below we briefly review SSGE. More details refer to \cite{shi2018spectral}.
Specifically, we denote  the target gradient function to estimate by $\mathbf{g} : \mathcal{X} \rightarrow \mathbb{R}^{d}:\mathbf{g}(\mathbf{x})=\nabla_{\mathbf{x}}\log q(\mathbf{x})$.
The $i^{th}$ component of the gradient is $g_{i}(\mathbf{x})=\nabla_{\mathbf{x}_{i}} \log q(\mathbf{x})$. 
We assume $g_{1}, \ldots, g_{d} \in L^{2}(\mathcal{X}, q)$. $\left\{\psi_{j}\right\}_{j \geq 1}$ denotes an orthonormal basis of $L^{2}(\mathcal{X}, q)$.
We can expand $g_{i}(\mathbf{x})$ into the spectral series, i.e., $g_{i}(\mathbf{x})=\sum_{j=1}^\infty\beta_{ij}\psi_j(\mathbf{x})$.
%\begin{align}
%\label{sps}
%g_{i}(\mathbf{x})=\sum_{j=1}^\infty\beta_{ij}\psi_j(\mathbf{x}).
%\end{align}
The value of the $j^{th}$ eigenfunction $\psi_{j}$ at $\mathbf{x}$ can be approximated by the Nystr$\ddot{o}$m method \citep{DBLP:conf/aaai/XuJSZ15}.
Due to the orthonormality of eigenfunctions $\left\{\psi_{j}\right\}_{j \geq 1}$,  there is a constraint under the probability measure q(.): $\int \psi_{i}(\mathbf{x}) \psi_{j}(\mathbf{x}) q(\mathbf{x}) d \mathbf{x}=\delta_{i j}$, where $\delta_{i j}=\mathbbm{1}[i=j]$.
Based on this constraint, we can obtain the following equation for $\left\{\psi_{j}\right\}_{j \geq 1}$:
\begin{align}\label{kereq}
\int k(\mathbf{x}, \mathbf{y}) \psi(\mathbf{y}) q(\mathbf{y}) d \mathbf{y}=\mu \psi(\mathbf{x}),
\end{align}
where $k(.)$ is a kernel function. 
The left side of the above equation can be approximated by the Monte Carlo estimate using i.i.d. samples {$\mathbf{x}^1, . . . , \mathbf{x}^M $} from $q(.)$ : $\frac{1}{M} \mathbf{K} \psi \approx \mu \psi$, where $\mathbf{K} $ is the Gram Matrix and $ \psi =\left[\psi\left(\mathbf{x}^{1}\right), \ldots, \psi\left(\mathbf{x}^{M}\right)\right]^{\top} $. We can solve this eigenvalue problem by choose the $J$ largest eigenvalues $\lambda_{1} \geq \cdots \geq \lambda_{J}$ for $\mathbf{K} $. $\boldsymbol{u}_j$ denotes  the eigenvector of the Gram matrix. The approximation for $\left\{\psi_{j}\right\}_{j \geq 1}$ can be obtained combined with Eq. (\ref{kereq}) as following:
$\psi_{j}(\mathbf{x}) \approx \hat{\psi}_{j}(\mathbf{x})=\frac{\sqrt{M}}{\lambda_{j}} \sum_{m=1}^{M} u_{j m} k\left(\mathbf{x}, \mathbf{x}^{m}\right)$.

% \begin{align}
% \psi_{j}(\mathbf{x}) \approx \hat{\psi}_{j}(\mathbf{x})=\frac{\sqrt{M}}{\lambda_{j}} \sum_{m=1}^{M} u_{j m} k\left(\mathbf{x}, \mathbf{x}^{m}\right),
% \end{align}

%Next, we aim to estimate the coefficients $\beta_{ij}$ by Stein’s identity. Substituting in Equation (\ref{sps}) into  Equation (\ref{steinid})  and using the orthonormality of $\left\{\psi_{j}\right\}_{j \geq 1}$ , we can obtain
Furthermore, based on the orthonormality of $\left\{\psi_{j}\right\}_{j \geq 1}$, we can easily obtain
$\beta_{i j}=-\mathbb{E}_{q} \nabla_{\mathbf{x}_{i}} \psi_{j}(\mathbf{x})$.
%We can further approximate $\nabla_{\mathbf{x}_{i}} \psi_{j}(\mathbf{x})$ to estimate $\beta_{ij}$. 
By taking derivative both sides of Eq. (\ref{kereq}), we can show that:
\begin{align} \mu_{j} \nabla_{\mathbf{x}_{i}} \psi_{j}(\mathbf{x}) =\nabla_{\mathbf{x}_{i}} \int k(\mathbf{x}, \mathbf{y}) \psi_{j}(\mathbf{y}) q(\mathbf{y}) d \mathbf{y} =\int \nabla_{\mathbf{x}_{i}} k(\mathbf{x}, \mathbf{y}) \psi_{j}(\mathbf{y}) q(\mathbf{y}) d \mathbf{y}. \end{align}
Then we can estimate as following:
\begin{align}
\hat{\nabla}_{\mathbf{x}_{i}} \psi_{j}(\mathbf{x}) \approx \frac{1}{\mu_{j} M} \sum_{m=1}^{M} \nabla_{\mathbf{x}_{i}} k\left(\mathbf{x}, \mathbf{x}^{m}\right) \psi_{j}\left(\mathbf{x}^{m}\right).
\end{align}
Finally, by truncating the expansion to the first $J$ terms and plugging in the Nystr$\ddot{o}$m approximations of $\left\{\psi_{j}\right\}_{j \geq 1}$, we can get the score estimator:

% $\hat{g}_{i}(\mathbf{x}) =\sum_{j=1}^{J} \hat{\beta}_{i j} \hat{\psi}_{j}(\mathbf{x}),~~~~~~ \hat{\beta}_{i j} =-\frac{1}{M} \sum_{m=1}^{M} \nabla_{\mathbf{x}_{i}} \hat{\psi}_{j}\left(\mathbf{x}^{m}\right)$

\begin{align}
\hat{g}_{i}(\mathbf{x}) =\sum_{j=1}^{J} \hat{\beta}_{i j} \hat{\psi}_{j}(\mathbf{x}),~~~~~~ \hat{\beta}_{i j} =-\frac{1}{M} \sum_{m=1}^{M} \nabla_{\mathbf{x}_{i}} \hat{\psi}_{j}\left(\mathbf{x}^{m}\right).
\end{align}

% where $\hat{\psi}_{j}(\mathbf{x})$ is the Nystr$\ddot{o}$m approximation of $ \psi_j(\mathbf{x})$.

% where $\hat{\psi}_{j}(\mathbf{x})$ is the Nystr$\ddot{o}$m approximation of $ \psi_j(\mathbf{x})$.

In general, representation learning for large-scale datasets is usually costly in terms of storage and computation. For instance, the dimension of images in the STL-10 dataset is $96\times 96\times3$ (i.e., the vector length is 27648). This makes it almost impossible to directly estimate the gradient of MI between the input and representation. To alleviate this problem, we introduce random projection (RP) ~\citep{DBLP:conf/kdd/BinghamM01} to reduce the dimension of $\mathbf{x}$. 

We  briefly review RP. More details refer to \cite{DBLP:conf/kdd/BinghamM01}. RP projects the original $d$-dimensional data into a $k$-dimensional $(k<<d)$ subspace. Concretely, let matrix $X_{d \times N}$ denotes the 
original set of N $d$-dimensional data, the projection of the original data $X_{k \times N}^{R P}$ is obtained by introducing a random matrix $R_{k \times d}$ whose columns have unit length, as follows~\citep{DBLP:conf/kdd/BinghamM01},
$X_{k \times N}^{R P}=R_{k \times d} X_{d \times N}.$
After  RP, the Euclidean distance between two original data vectors can be approximated by the Euclidean distance of the projective vectors in reduced spaces:
\begin{align}\label{rj}
\left\|\mathbf{x}_{1}-\mathbf{x}_{2}\right\| \approx \sqrt{d / k}\left\|R \mathbf{x}_{1}-R \mathbf{x}_{2}\right\|,
\end{align}
where $\mathbf{x}_{1}$ and $\mathbf{x}_{2}$ denote the two data vectors in the original large dimensional space.

Based on the principle of RP, we can derive a Salable Spectral Stein Gradient Estimator, which is an efficient high-dimensional score function estimator. One can show that the RBF kernel satisfies Stein’s identity~\citep{liu2016stein}.
\cite{shi2018spectral} also shows that it is a promising choice for SSGE with a lower error bound. %A major computational bottleneck lies in the computation of the gram matrix.
%Its formulation is as following: $k\left(\mathbf{x}_1, \mathbf{x}_2\right)=\exp \left(-\left\|\mathbf{x}_1-\mathbf{x}_2\right\|^{2} / 2 \sigma^{2}\right)$, where $\sigma$ is the bandwidth.
%To reduce the computation of $k\left(\mathbf{x}_1, \mathbf{x}_2\right)$ of SSGE in high-dimensional settings, we replace the input of SSGE with a projections obtained by RP according to the appropriation of Eq. (\ref{rj}) for the computation of RBF kernel. 
To reduce the computation of the kernel similarities of SSGE in high-dimensional settings, we replace the input of SSGE with a projections obtained by RP according to the approximation of Eq. (\ref{rj}) for the computation of the RBF kernel.

\section{Mutual Information Gradient Estimator}

%%%%%%%%%%added by lirong
As gradient estimation is a straightforward and effective method in optimization, we  propose a gradient estimator for MI based on score estimation of implicit distributions, which is called Mutual Information Gradient estimator (MIGE). In this section, we focus on three most general cases of MI gradient estimation for representation learning, and derive the corresponding MI gradient estimator for these circumstances.

We outline the general setting of training an encoder to learn a representation. Let $\mathcal{X}$  and $\mathcal{Z}$ be the domain, and $E_{\psi}: \mathcal{X} \rightarrow \mathcal{Z}$ with parameters $\psi$ denotes a continuous and (almost everywhere) differentiable parametric function, which is usually a neural network, namely an encoder. $p(\mathbf{x})$ denotes the empirical distribution given the input data $\mathbf{x}  \in \mathcal{X}$. We  can obtain  the representation of the input data through the encoder, $\mathbf{z}= E_{\psi}(\mathbf{x})$. $q_{\psi}(\mathbf{z})$ is defined as the marginal distribution induced by pushing samples from $p(\mathbf{x})$ through encoder  $E_{\psi}(.)$ We also define $q_\psi(\mathbf{x},\mathbf{z})$ as the joint distribution with $\mathbf{x}$ and $\mathbf{z}$, which is determined by encoder $E_{\psi}(.)$.

\textbf{Circumstance \uppercase\expandafter{\romannumeral1}}.~~Given that the encoder $E_{\psi}(.)$ is deterministic, our goal is to estimate the gradient of MI between input $\mathbf{x}$ and encoder output $\mathbf{z}$  w.r.t. the encoder parameters $\psi$. There is a close relationship between mutual information and entropy, which is as following:$I_\psi(\mathbf{x};\mathbf{z})=H(\mathbf{x})+H_\psi(\mathbf{z})-H_\psi(\mathbf{x},\mathbf{z})$. Here $H(\mathbf{x})$ is data entropy and not relevant to $\psi$. The optimization of $I_\psi(\mathbf{x},\mathbf{z})$ with parameters $\psi$ can neglect the entry $H(\mathbf{x})$. We decompose the gradient of the entropy of $q_{\psi}(\mathbf{z})$ and $q_{\psi}(\mathbf{x},\mathbf{z})$  as (see Appendix \ref{A_entropy}):
 \begin{align}
 \nabla_{\psi} H(\mathbf{z})  
   =-\nabla_{\psi} \mathbb{E}_{ q_{\psi}(\mathbf{z})} [\log q(\mathbf{z})], ~~~~
   \nabla_{\psi} H(\mathbf{x},\mathbf{z})  
   =-\nabla_{\psi} \mathbb{E}_{ q_{\psi}(\mathbf{x},\mathbf{z})} [\log q(\mathbf{x},\mathbf{z})].
\end{align}
Hence, we can represent the gradient of MI between input $\mathbf{x}$ and encoder output $\mathbf{z}$ w.r.t. encoder parameters $\psi$ as following:
\begin{align}\label{mige1}
\nabla_{\psi} I_\psi(\mathbf{x};\mathbf{z})=-\nabla_{\psi} \mathbb{E}_{ q_{\psi}(\mathbf{z})} [\log q(\mathbf{z})]+\nabla_{\psi} \mathbb{E}_{ q_{\psi}(\mathbf{x},\mathbf{z})} [\log q(\mathbf{x},\mathbf{z})].
\end{align}
However, this equation is intractable since an expectation w.r.t $q_\psi(\mathbf{z})$ is directly not differentiable  w.r.t $\psi$. ~\cite{roeder2017sticking} proposed a general variant of the
standard reparameterization trick for the variational evidence lower bound, which demonstrates lower-variance. 
To address above problem, we adapt this trick for MI gradient estimator in representation learning.
Specifically, we can obtain the samples from the marginal distribution of $\mathbf{z}$ by pushing samples from the data empirical distribution  $p(\mathbf{x})$ through  $E_{\psi}(.)$ for representation learning.  
Hence we can reparameterize the representations variable $\mathbf{z} \sim q_{\psi}(\mathbf{z})$ using a differentiable transformation:$ \mathbf{z}=E_{\psi}(\mathbf{x}) ~~~~~~\text{with} ~~~~~~\mathbf{x}~\sim  ~p(\mathbf{x})$, where the data empirical distribution $p(\mathbf{x})$ is independent of encoder parameters $\psi$. This reparameterization can rewrite an expectation w.r.t $q_\psi(\mathbf{z})$ and $q_\psi(\mathbf{x},\mathbf{z})$ such that the Monte Carlo estimate of the expectation is differentiable w.r.t $\psi$.

Relying on this reparameterization trick, we can represent  the gradient of MI w.r.t. encoder parameters $\psi$ in Eq. \ref{mige1} as follows:
\begin{align}
\nabla_{\psi} I_\psi(\mathbf{x};\mathbf{z})
   &=-\mathbb{E}_{ q(\mathbf{x})} [\nabla_{\mathbf{z}}\log q(E_{\psi}(\mathbf{x}))\nabla_{\psi}E_{\psi}(\mathbf{x})]\nonumber\\
   &+\mathbb{E}_{ q(\mathbf{x})} [\nabla_{(\mathbf{x},\mathbf{z})}\log q(\mathbf{x},E_{\psi}(\mathbf{x}))\nabla_{\psi}(\mathbf{x},E_{\psi}(\mathbf{x}))],
\end{align}
where the score function $\nabla_{\mathbf{z}} \log q_{\boldsymbol{\psi}}(E_{\psi}(\mathbf{x}))$ can be estimated based on i.i.d. samples from an implicit density $q_{\psi}(\mathbf{E_{\psi}(\mathbf{x})})$~\citep{shi2018spectral,song2019sliced}.
The samples form the joint distribution $q_{\psi}(\mathbf{x},\mathbf{z})$ are produced as following: we sample observations from empirical distribution $p(\mathbf{x})$; then the corresponding samples of $\mathbf{z}$ is obtained through $E_{\psi}(.)$. Hence we can also estimate $\nabla_{(\mathbf{x},\mathbf{z})}\log q(\mathbf{x},E_{\psi}(\mathbf{x}))$ based on i.i.d. samples from $q_{\psi}(\mathbf{x},E_{\psi}(\mathbf{x}))$. 
$\nabla_{\psi}E_{\psi}(\mathbf{x})$ and $\nabla_{\psi}(\mathbf{x},E_{\psi}(\mathbf{x}))$ are directly computed with $\mathbf{x}$.

\textbf{Circumstance \uppercase\expandafter{\romannumeral2}}.~~ Assume that we encode the input to latent data space $\mathbf{h}=C_{\psi}(\mathbf{x})$ that reflects useful structure in the data. Next, we summarize this latent variable mapping into final representations by the function $f_{\psi}$, $z=E_{\psi}(\mathbf{x})=f_{\psi} \circ C_{\psi}(\mathbf{x})$. The gradient 
estimator of MI between $\mathbf{h}$ and $\mathbf{z}$ is represented by the data reparameterization trick as follows:
\begin{align}
\nabla_{\psi} I_\psi(\mathbf{h};\mathbf{z})&= \nabla_{\psi} H_\psi(\mathbf{h})+ \nabla_{\psi} H_\psi(\mathbf{z})- \nabla_{\psi} H_\psi(\mathbf{h},\mathbf{z})\\
 &=-\mathbb{E}_{ q(\mathbf{x})} [\nabla_{\mathbf{z}}\log q(E_{\psi}(\mathbf{x}))\nabla_{\psi}E_{\psi}(\mathbf{x})]\nonumber-\mathbb{E}_{ q(\mathbf{x})} [\nabla_{\mathbf{h}}\log q(C_{\psi}(\mathbf{x}))\nabla_{\psi}C_{\psi}(\mathbf{x})]\nonumber\\
   &+\mathbb{E}_{ q(\mathbf{x})} [\nabla_{(\mathbf{h},\mathbf{z})}\log q(C_\psi(\mathbf{x}),E_{\psi}(\mathbf{x}))\nabla_{\psi}(C_\psi\mathbf{x},E_{\psi}(\mathbf{x}))].
\end{align}
\textbf{Circumstance \uppercase\expandafter{\romannumeral3}}.~~ Consider stochastic encoder function $E_{\psi}(.,\boldsymbol{\epsilon})$ where $\boldsymbol{\epsilon}$ is an auxiliary variable with independent marginal $p(\boldsymbol{\epsilon})$. By utilizing data reparameterization trick.  we can represent the gradient of the conditional entropy $H_{\psi}(\mathbf{z}|\mathbf{x})$ as follows (see Appendix A):
\begin{align}\label{centropy}
 \nabla_{\psi} H_{\psi}(\mathbf{z}|\mathbf{x})=-\mathbb{E}_{p(\mathbf{x})}[\mathbb{E}_{p(\boldsymbol{\epsilon})}[\nabla_{(\mathbf{z}|\mathbf{x})}\log q(E_\psi(\mathbf{x},\boldsymbol{\epsilon})|\mathbf{x})\nabla_{\psi}E_\psi(\mathbf{x},\boldsymbol{\epsilon})]],
\end{align}
where the term  $\nabla_{(\mathbf{z}|\mathbf{x})}\log q(E_\psi(\mathbf{x},\boldsymbol{\epsilon})|\mathbf{x})$ can be easily estimated by score estimation. 

Based on the condition entropy  gradient estimation in Eq. (\ref{centropy}), the gradient estimator of MI between input and encoder output can be represented as following: 
 \begin{align}
 \nabla_{\psi} I_\psi(\mathbf{x};\mathbf{z})&=\nabla_{\psi} H_\psi(\mathbf{z})-\nabla_{\psi} H_\psi(\mathbf{z|x}) \\
 &=-\mathbb{E}_{ p(\mathbf{x})p(\boldsymbol{\epsilon})} [\nabla_{\mathbf{z}}[\log p(E_{\psi}(\mathbf{x},\boldsymbol{\epsilon}))]\nabla_{\psi}E_{\psi}(\mathbf{x},\boldsymbol{\epsilon})]\nonumber\\
 &+\mathbb{E}_{p(\mathbf{x})}[\mathbb{E}_{p(\boldsymbol{\epsilon})}[\nabla_{(\mathbf{z}|\mathbf{x})}\log q(E_\psi(\mathbf{x},\boldsymbol{\epsilon})|\mathbf{x})\nabla_{\psi}E_\psi(\mathbf{x},\boldsymbol{\epsilon})]].
\end{align}
In practical MI optimization, we can construct MIGE of the full dataset based on mini-batch Monte Carlo estimates. We have provided an algorithm description for MIGE in Appendix B.

% % The bias of MIGE depends on score estimation of implicit distributions. We prefer Spectral Stein Gradient Estimator (SSGE) for score estimation. 

% %  And the error bound of SSGE is proved in~\cite{shi2018spectral}.

\section{Toy Experiment}
\label{sec:toy}
Recently, MINE and MINE-$f$ enable effective computation of MI in the continuous and high-dimensional settings. To compare with MINE and MINE-$f$, we evaluate MIGE in the correlated Gaussian problem taken from ~\citep{belghazi2018mine}.

\textbf{Experimental Settings}. We consider two random variables $\mathbf{x}$ and $\mathbf{y}$ ($\mathbf{x},\mathbf{y}\in\mathcal{R}^{d}$), coming from a $2d$-dimension multivariate Gaussian distribution. The component-wise correlation of $\mathbf{x}$ and $\mathbf{y}$ is defined as follows:
$corr(\mathbf{x}_i,\mathbf{y}_i)=\delta_{ij}\rho,~~\rho\in (-1,1)$,
where $\delta_{ij}$ is Kronecker's delta and $\rho$ is the correlation coefficient. Since MI is invariant to smooth transformations of $\mathbf{x, y}$, we only consider standardized Gaussian for marginal distribution $p(\mathbf{x})$ and $p(\mathbf{y})$. The gradient of MI w.r.t $\rho$ has the analytical solution: $\nabla_{\rho} I(\mathbf{x;y}) = \frac{\rho d}{1 - \rho^2}$.
We apply MINE and MINE-$f$ to estimate MI of $\mathbf{x,y}$ by sampling from the correlated Gaussian distribution and its marginal distributions, and the corresponding gradient of MI  w.r.t $\rho$ can be computed by backpropagation implemented in Pytorch.
%MIGE receives 256 samples drawn from the correlated Gaussian distribution and its marginal distribution. In contrast, MINE and MINE-$f$ receive 51200 samples (256 samples $\times$ 200 epochs) since the DNN estimators they rely on need a large number of samples to train.

\textbf{Results}. Fig.\ref{fig:toy} presents our experimental results in different dimensions $d=\{5, 10, 20\}$. In the case of low-dimensional $(d=5)$, all the estimators give promising estimation of MI and its gradient. However, the MI estimation of MINE and MINE-$f$ are unstable due to its relying on a discriminator to produce estimation of the bound on MI. Hence, as showed in Fig.\ref{fig:toy}, corresponding estimation of MI and its gradient is not smooth. As the dimension $d$ and the absolute value of correlation coefficient $\left|\rho \right|$ increase, MINE and MINE-$f$ are apparently hard to reach the True MI, and their gradient estimation of MI is thus high biased. This phenomenon would be more significant in the case of high-dimensional or large MI. Contrastively, MIGE demonstrates  the  significant improvement over MINE and MINE-$f$ when estimating MI gradient between twenty-dimensional random variables $\mathbf{x, y}$.
In this experiment, we compare our method with two baselines on an analyzable problem and find that the gradient curve estimated by our method is far superior to other methods in terms of smoothness and tightness in a high-dimensional and large-MI setting compared with MINE and MINE-$f$.

% Fig.\ref{fig:toy} shows our results in different dimensions $d=\{5, 10, 20\}$.
% In the case of low-dimension $(d=5)$, all methods give satisfactory gradient estimation, although the curves of MINE and MINE-$f$ are not so steady. In the case of high-dimension $(d=20)$, MINE and MINE-$f$ are apparently hard to reach the True MI, and their gradient estimation is thus high biased. Whereas in any dimension, our approach gives rather good estimation.

% The unsteadiness of MINE and MINE-$f$ originates from the instability of neural networks across training process, and would transmit into downwards derivation, as we observed, finally causes corresponding gradient estimation relatively unsmooth.
% The looseness of the variational lower bounds used by MINE and MINE-$f$ would result in gradient decay. As the dimension increases, the mutual information increases accordingly, and this phenomenon would be more significant. However, our approach overcomes these shortcomings and always gives good estimation with excellent property: steadiness, smoothness and accuracy.

\section{Applications}
\label{sec:app}
To demonstrate the performance in downstream tasks, we deploy MIGE to Deep InfoMax \citep{DBLP:conf/iclr/HjelmFLGBTB19} and Information Bottleneck \citep{tishby2000information} respectively, namely replacing the original MI estimators with MIGE. We find that MIGE achieves higher and more stable classification accuracy, which indicating its good gradient estimation performance in practical applications.
% In our experiments, we use the Stein gradient estimator~\citep{shi2018spectral} to estimate the score function term.

\subsection{Deep InfoMax} 
Discovering useful representations from unlabeled data is one core problem for deep learning. Recently, a growing set of methods is explored to train deep neural network encoders by maximizing the mutual information between its input and output.  
A number of methods based on tractable variational lower bounds, such as JSD and infoNCE, have been proposed to improve the estimation of MI between high dimensional input/output pairs of deep neural networks \citep{DBLP:conf/iclr/HjelmFLGBTB19}.
To compare with JSD and infoNCE, we expand the application of MIGE in unsupervised learning of deep representations based on the InfoMax principle.

\textbf{Experimental Settings.}   
For consistent comparison, we follow the experiments of Deep InfoMax(DIM)\footnote{Codes available at \url{https://github.com/rdevon/DIM}} to set the experimental setup as in \cite{DBLP:conf/iclr/HjelmFLGBTB19}. 
We test DIM on image datasets CIFAR-10, CIFAR-100 and STL-10 to evaluate our MIGE.
For the high-dimensional images in STL-10, directly applying SSGE is almost impossible since it results in exorbitant computational cost. Our proposed Scalable SSGE is applied, to reduce the dimension of images and achieve reasonable computational cost.
As mentioned in \cite{DBLP:conf/iclr/HjelmFLGBTB19}, non-linear classifier is chosen to evaluate our representation,
After learning representation, we freeze the parameters of the encoder and train a non-linear classifier using the representation as the input. The same classifiers are used for all methods. Our baseline results are directly copied from \cite{DBLP:conf/iclr/HjelmFLGBTB19} or by running the code of author.

\begin{table}[t]

\caption{CIFAR-10 and CIFAR-100 classification accuracy (top 1) of downstream tasks compared with vanilla DIM. JSD and infoNCE are MI estimators, and PM denotes matching representations to a prior distribution \citep{DBLP:conf/iclr/HjelmFLGBTB19}.}
    \label{table:infomax}
    \begin{center}
    \begin{tabular}{c||ccc|ccc}
    \multirow{2}{*}{\textbf{Model}} & \multicolumn{3}{c|}{\textbf{CIFAR-10}} & \multicolumn{3}{c}{\textbf{CIFAR-100}} \\ 
                        & conv           & fc(1024)       & Y(64)          & conv           & fc(1024)       & Y(64) \\ \hline 
    % Fully supervised    & \multicolumn{3}{c|}{75.39}                      & \multicolumn{3}{c}{42.27}               \\ \hline
    DIM (JSD)                 & 55.81\%        & 45.73\%       & 40.67\%          & 28.41\%          & 22.16\%          & 16.50\% \\
    DIM (JSD + PM)            & 52.2\%           & 52.84\%          & 43.17\%          & 24.40\%          & 18.22\%          & 15.22\% \\
    DIM (infoNCE)             & 51.82\%          & 42.81\%          & 37.79\%          & 24.60\%          & 16.54\%          & 12.96\% \\
    DIM (infoNCE + PM)        & 56.77\%          & 49.42\%          & 42.68\%          & 25.51\%          & 20.15\%          & 15.35\% \\ \hline
    MIGE             & \textbf{57.95\%} & \textbf{57.09\%} & \textbf{53.75\%} & \textbf{29.86\%} & \textbf{27.91\%} & \textbf{25.84\%} \\ \hline
    \end{tabular}
    \end{center}
\end{table}

%CIFAR-10 and CIFAR-100 each consists of  32$\times$32  colored images, with 50,000 training images and 10,000 testing images.  
%We adopt the same encoder architecture used in~\cite{DBLP:conf/iclr/HjelmFLGBTB19}, which uses a deep convolutional GAN (DCGAN, \cite{radford2015unsupervised}) consisting of 3 convolutional layers and 2 fully connected layer. The same empirical setup is used. Follow \cite{DBLP:conf/iclr/HjelmFLGBTB19}, we choose image classification as the downstream task, then evaluate our representation in terms of the accuracy of transfer learning classification, that is, freezing the weights of the encoder and training a small fully-connected neural network classifier using the representation as the input.

\begin{figure}
\begin{minipage}{0.6\linewidth}
\tabcaption{STL-10 classification accuracy (top 1) of downstream tasks compared with vanilla DIM. The dimension of STL-10 images (27648) results in exorbitant computational cost. Random Projection (RP) is applied to reduce the dimension.}
    \begin{center}
    \label{table:infomax-STL}
    \begin{tabular}{c||c c c}
    \multirow{2}{*}{\textbf{Model}} & \multicolumn{3}{c}{\textbf{STL-10}} \\ 
                        & conv           & fc(1024)       & Y(64)          \\ \hline 
    DIM (JSD)           & 42.03\%          & 30.28\%         & 28.09\%          \\
    DIM (infoNCE)       & 43.13\%          & 35.80\%          & 34.44\%          \\ \hline
    MIGE                & \multicolumn{3}{c}{unaffordable computational cost}    \\
    % MIGE + RP to 1024d  & 51.43\% & 47.24\% & \textbf{45.28\%} \\
    MIGE + RP to 512d   & 52.00\% & 48.14\% & 44.89\% \\
    %MIGE + RP to 256d   & 51.01\% & 47.30\% & 44.75\% \\
    %MIGE + RP to 128d   & 50.90\% & 47.51\% & 44.96\% \\ 
    \hline
    \end{tabular}
    \end{center}
\end{minipage}
\begin{minipage}{0.4\linewidth}
    \centering
    \includegraphics[width=2.1in, keepaspectratio]{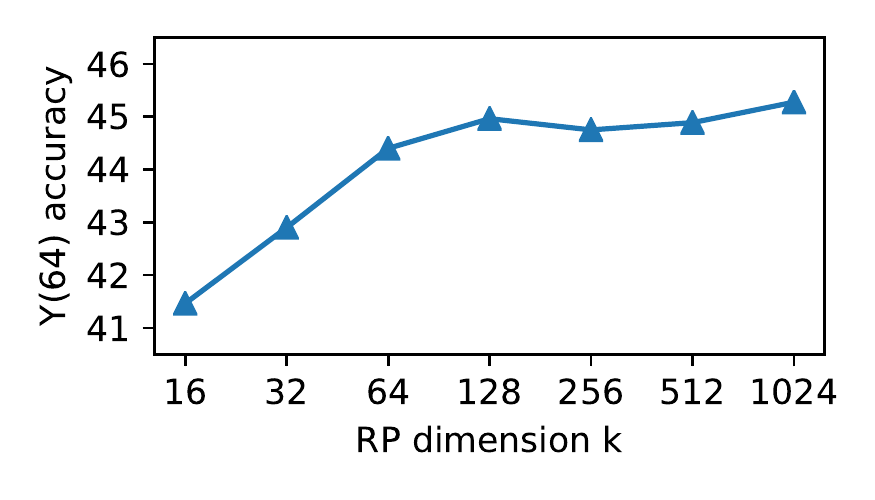}
    \caption{STL-10 Y(64) classification accuracy (top 1) with different RP dimension.}
    \label{fig:rp}
\end{minipage}
\end{figure}

\textbf{Results.} As shown in Table \ref{table:infomax},  MIGE outperforms all the competitive models in DIM experiments on CIFAR-10 and CIFAR-100.
Besides the numerical improvements, it is notable that our model have the less accuracy decrease across layers than that of DIM(JSD) and DIM(infoNCE). The results indicate that, compared to variational lower bound methods, MIGE gives more favorable gradient direction, and demonstrates more power in controlling information flows without significant loss.
With the aid of Random Projection, we could evaluate on bigger datasets, e.g., STL-10.
Table \ref{table:infomax-STL}  shows the result of DIM experiments on STL-10.
We can observe significant improvement over the baselines when RP to 512d.  Note that our proposed gradient estimator can also be extended to the multi-view setting(i.e., with local and global features) of DIM, it is beyond the scope of this paper. More discussions refer to Appendix C.
 %[t]

\textbf{Ablation Study.} 
To verify the effect of different dimensions of Random Projection on classification accuracy in DIM experiments, we conduct an  ablation study on STL-10 with the above experimental settings. Varying RP dimension $k \in\{16,32,64,128,256,512,1024\}$, we measure the classification accuracy of Y(64)  which is shown in Fig.\ref{fig:rp}. We find that the classification accuracy increases with RP dimension from 16 to 128. %The further increase of RP dimension generally did not significantly improve the classification accuracy. %Intuitively, the  RP dimension is closely related to the model capacity. Hence the increase of RP dimension from 16 to 128 can greatly improve the performance of MIGE.
After that, the approximation in Equ.(\ref{rj}) with the further increase of the RP dimension reaches saturation, while bringing extra computational costs. 

% \begin{table}[t]
% \caption{InfoMax on CIFAR10}
% \label{cifar10}
% \begin{center}
% \begin{tabular}{cccc}
% \textbf{Model}      & \textbf{conv}  & \textbf{fc(1024)}  & \textbf{Y(64)} \\ \hline 
% Fully supervised    & \multicolumn{3}{c}{75.39}                            \\ \hline
% JSD                 & 55.81          & 45.73              & 40.67          \\
% JSD + PM            & 52.2           & 52.84              & 43.17          \\
% infoNCE             & 51.82          & 42.81              & 37.79          \\
% infoNCE + PM        & 56.77          & 49.42              & 42.68          \\ \hline
% SP (ours)           & \textbf{57.95} & \textbf{57.09}     & \textbf{53.75} \\ \hline
% \end{tabular}
% \end{center}
% \end{table}

% \begin{table}[t]
% \caption{InfoMax on CIFAR100}
% \label{cifar100}
% \begin{center}
% \begin{tabular}{cccc}
% \textbf{Model}      & \textbf{conv}  & \textbf{fc(1024)}  & \textbf{Y(64)} \\ \hline 
% Fully supervised    & \multicolumn{3}{c}{42.27}                            \\ \hline
% JSD                 & 28.41          & 22.16              & 16.50          \\
% JSD + PM            & 24.40          & 18.22              & 15.22          \\ 
% infoNCE             & 24.60          & 16.54              & 12.96          \\
% infoNCE + PM        & 25.51          & 20.15              & 15.35          \\ \hline
% MIGE (ours)           & \textbf{29.86} & \textbf{27.91}     & \textbf{25.84} \\ \hline
% \end{tabular}
% \end{center}
% \end{table}

\subsection{Information Bottleneck}
Information Bottleneck (IB) has been widely applied to a variety of application domains, such as classification \citep{tishby2015deep,DBLP:conf/iclr/AlemiFD017,chalk2016relevant,kolchinsky2017nonlinear}, clustering~\citep{slonim2000document}, and coding theory and quantization~\citep{zeitler2008design,courtade2011multiterminal}. %IB is first introduced by \cite{Tishby1999IB} as a method of seeking a representation that weighed the sufficiency for the target and the complexity of the representation. 
In particular, given the input variable $\mathbf{x}$ and the target variable $\mathbf{y}$, the goal of the IB is to learn a representation of $\mathbf{x}$ (denoted by the variable $\mathbf{z}$) that satisfies the following characteristics:
\begin{itemize}
\item[1)] $\mathbf{z}$ is sufficient for the target $\mathbf{y}$, that is, all information about target $\mathbf{y}$ contained in $\mathbf{x}$ should also be contained in $\mathbf{z}$. In optimization, it should be %achieved by maximizing the information between $\mathbf{y}$ and $\mathbf{z}$.
\item[2)] $\mathbf{z}$ is minimal. %It can be known that there are many representations satisfying the point 1). 
In order not to contain irrelevant information that is not related to $\mathbf{y}$, $\mathbf{z}$ is required to contain the smallest information among all sufficient representations. 
\end{itemize}
%Since mutual information quantifies the dependence between two random   variables, IB introduces it to characterize the above two characteristics. The first characteristic above can be represented by $I(\mathbf{z};\mathbf{y})=I(\mathbf{z};\mathbf{x})$. In detail, we implement this by maximizing the $I(\mathbf{z};\mathbf{y})$. And the second characteristic above indicates that $I(\mathbf{z};\mathbf{x})$ should be smallest among all possible representations. More specifically, the IB applies a natural constraint to implement the second point, namely $I(\mathbf{z},\mathbf{x}) \leq C$ (\cite{witsenhausen1975conditional}), where $c$ is the information constraint.
The objective function of IB is written as follows:
\begin{align}
\max I(\mathbf{z};\mathbf{y}), \hspace{2ex} \text{s.t.} \hspace{1ex} I(\mathbf{z};\mathbf{x})\leq c.
\end{align}
Equivalently, by introducing a Lagrangian multiplier $\beta$, the IB method can maximize the following objective function:
$G_{IB} =I(\mathbf{z};\mathbf{y})-\beta I(\mathbf{z};\mathbf{x}).$
Further, it is generally acknowledged that $I(\mathbf{z};\mathbf{y})=H(\mathbf{y})-H(\mathbf{y}|\mathbf{z})$, and $H(\mathbf{y})$ is constant. Hence we can also minimize the objective function of the following form:
\begin{align}
    L_{IB}=H(\mathbf{y}|\mathbf{z})+\beta I(\mathbf{z};\mathbf{x}),
    \label{IBloss}
\end{align}
where $\beta \geq 0$ plays a role in trading off the sufficiency and minimality. Note that the above formulas omit the parameters for simplicity. 

To overcome the intractability of MI in the continuous and high-dimension setting,  \cite{DBLP:conf/iclr/AlemiFD017} presents a variational approximation to IB, which adopts deep neural network encoder to produce a conditional multivariate normal distribution, called Deep Variational Bottleneck (DVB). Rencently, DVB is exploited to restrict the capacity of discriminators in GANs  ~\citep{DBLP:conf/iclr/PengKTAL19}. However,
a tractable density is required for the approximate posterior in DVB due to their reliance on a variational approximation while MIGE does not. 

To evaluate our method, we compare MIGE-IB with DVB and MINE-IB in IB application. We demonstrate an implementation of the IB objective on permutation invariant MNIST using MIGE. 

\textbf{Experiments.}  For consistent comparison, we adopt the same architecture and empirical settings used in \cite{DBLP:conf/iclr/AlemiFD017} except that the initial learning rate of 2e-4 is set for Adam optimizer, and exponential decay with decaying rate by a factor of 0.96 was set for every 2 epochs. The implementation of DVB is available from its authors\footnote{\url{https://github.com/alexalemi/vib_demo}}. Under these experimental settings, we use our MI Gradient Estimator to replace the MI estimator in DVB experiment. 
The threshold of score function's Stein gradient estimator is set as $0.94$. The threshold is the hyper-parameter of Spectral Stein Gradient Estimator (SSGE), and it is used to set the kernel bandwidth of RBF kernel. Our results can be seen in Table \ref{table:ib} and it manifests that our proposed MIGE-IB outperforms DVB and MINE-IB.
\begin{table}[H]
\caption{Permutation-invariant MNIST misclassification rate. Datas except our model are cited from \cite{belghazi2018mine}}
\label{table:ib}
\begin{center}
\begin{tabular}{cc}
\textbf{Model}         & \textbf{Misclass rate} \\ \hline 
Baseline               & 1.38\%          \\
Dropout                & 1.34\%          \\
Confidence penalty     & 1.36\%          \\
Label Smoothing        & 1.4\%           \\
DVB                    & 1.13\%          \\
MINE-IB                   & 1.11\%          \\ \hline
MIGE-IB (ours)          & \textbf{1.05\%} \\ \hline
\end{tabular}
\end{center}
\end{table}

\section{Conclusion}
In this paper, we present a gradient estimator, called Mutual Information Gradient Estimator (MIGE), to avoid the various problems met in direct mutual information estimation. We manifest the effectiveness of gradient estimation of MI over direct MI estimation by applying it in unsupervised or supervised representation learning. %MIGE is applied to the InfoMax principle and the Information Bottleneck respectively, namely replacing the original mutual information estimation term. 
Experimental results have indicated the remarkable improvement over MI estimation in the Deep InfoMax method and the Information Bottleneck method.

\section*{Accknowledgement}
This work was partially funded by the National Key R\&D Program of China (No. 2018YFB1005100 \& No. 2018YFB1005104). 
\newpage
\bibliography{iclr2020_conference}

\begin{thebibliography}{3}
\providecommand{\natexlab}[1]{#1}
\providecommand{\url}[1]{\texttt{#1}}
\expandafter\ifx\csname urlstyle\endcsname\relax
  \providecommand{\doi}[1]{doi: #1}\else
  \providecommand{\doi}{doi: \begingroup \urlstyle{rm}\Url}\fi

\bibitem[Bengio \& LeCun(2007)Bengio and LeCun]{Bengio+chapter2007}
Yoshua Bengio and Yann LeCun.
\newblock Scaling learning algorithms towards {AI}.
\newblock In \emph{Large Scale Kernel Machines}. MIT Press, 2007.

\bibitem[Goodfellow et~al.(2016)Goodfellow, Bengio, Courville, and
  Bengio]{goodfellow2016deep}
Ian Goodfellow, Yoshua Bengio, Aaron Courville, and Yoshua Bengio.
\newblock \emph{Deep learning}, volume~1.
\newblock MIT Press, 2016.

\bibitem[Hinton et~al.(2006)Hinton, Osindero, and Teh]{Hinton06}
Geoffrey~E. Hinton, Simon Osindero, and Yee~Whye Teh.
\newblock A fast learning algorithm for deep belief nets.
\newblock \emph{Neural Computation}, 18:\penalty0 1527--1554, 2006.

\end{thebibliography}


\begin{thebibliography}{51}
\providecommand{\natexlab}[1]{#1}
\providecommand{\url}[1]{\texttt{#1}}
\expandafter\ifx\csname urlstyle\endcsname\relax
  \providecommand{\doi}[1]{doi: #1}\else
  \providecommand{\doi}{doi: \begingroup \urlstyle{rm}\Url}\fi

\bibitem[Alemi et~al.(2017)Alemi, Fischer, Dillon, and
  Murphy]{DBLP:conf/iclr/AlemiFD017}
Alexander~A. Alemi, Ian Fischer, Joshua~V. Dillon, and Kevin Murphy.
\newblock Deep variational information bottleneck.
\newblock In \emph{ICLR}, 2017.

\bibitem[Alemi et~al.(2018{\natexlab{a}})Alemi, Fischer, and
  Dillon]{alemi2018uncertainty}
Alexander~A Alemi, Ian Fischer, and Joshua~V Dillon.
\newblock Uncertainty in the variational information bottleneck.
\newblock \emph{arXiv preprint arXiv:1807.00906}, 2018{\natexlab{a}}.

\bibitem[Alemi et~al.(2018{\natexlab{b}})Alemi, Poole, Fischer, Dillon,
  Saurous, and Murphy]{DBLP:conf/icml/AlemiPFDS018}
Alexander~A. Alemi, Ben Poole, Ian Fischer, Joshua~V. Dillon, Rif~A. Saurous,
  and Kevin Murphy.
\newblock Fixing a broken {ELBO}.
\newblock In \emph{ICML}, 2018{\natexlab{b}}.

\bibitem[Barber \& Agakov(2003)Barber and Agakov]{barber2003algorithm}
David Barber and Felix~V Agakov.
\newblock The im algorithm: a variational approach to information maximization.
\newblock In \emph{Advances in neural information processing systems}, pp.\
  None, 2003.

\bibitem[Belghazi et~al.(2018)Belghazi, Baratin, Rajeswar, Ozair, Bengio,
  Courville, and Hjelm]{belghazi2018mine}
Mohamed~Ishmael Belghazi, Aristide Baratin, Sai Rajeswar, Sherjil Ozair, Yoshua
  Bengio, Aaron Courville, and R~Devon Hjelm.
\newblock Mine: mutual information neural estimation.
\newblock \emph{ICML}, 2018.

\bibitem[Bingham \& Mannila(2001)Bingham and Mannila]{DBLP:conf/kdd/BinghamM01}
Ella Bingham and Heikki Mannila.
\newblock Random projection in dimensionality reduction: applications to image
  and text data.
\newblock In \emph{Proceedings of the seventh {ACM} {SIGKDD} international
  conference on Knowledge discovery and data mining, San Francisco, CA, USA},
  pp.\  245--250, 2001.

\bibitem[Butte \& Kohane(1999)Butte and Kohane]{butte1999mutual}
Atul~J Butte and Isaac~S Kohane.
\newblock Mutual information relevance networks: functional genomic clustering
  using pairwise entropy measurements.
\newblock In \emph{Biocomputing 2000}, pp.\  418--429. World Scientific, 1999.

\bibitem[Chalk et~al.(2016)Chalk, Marre, and Tkacik]{chalk2016relevant}
Matthew Chalk, Olivier Marre, and Gasper Tkacik.
\newblock Relevant sparse codes with variational information bottleneck.
\newblock In \emph{Advances in Neural Information Processing Systems}, pp.\
  1957--1965, 2016.

\bibitem[Courtade \& Wesel(2011)Courtade and Wesel]{courtade2011multiterminal}
Thomas~A Courtade and Richard~D Wesel.
\newblock Multiterminal source coding with an entropy-based distortion measure.
\newblock In \emph{2011 IEEE International Symposium on Information Theory
  Proceedings}, pp.\  2040--2044. IEEE, 2011.

\bibitem[Darbellay \& Vajda(1999)Darbellay and Vajda]{darbellay1999estimation}
Georges~A Darbellay and Igor Vajda.
\newblock Estimation of the information by an adaptive partitioning of the
  observation space.
\newblock \emph{IEEE Transactions on Information Theory}, 45\penalty0
  (4):\penalty0 1315--1321, 1999.

\bibitem[Fleuret(2004)]{fleuret2004fast}
Fran{\c{c}}ois Fleuret.
\newblock Fast binary feature selection with conditional mutual information.
\newblock \emph{Journal of Machine learning research}, 5\penalty0
  (Nov):\penalty0 1531--1555, 2004.

\bibitem[Fraser \& Swinney(1986)Fraser and Swinney]{fraser1986independent}
Andrew~M Fraser and Harry~L Swinney.
\newblock Independent coordinates for strange attractors from mutual
  information.
\newblock \emph{Physical review A}, 33\penalty0 (2):\penalty0 1134, 1986.

\bibitem[Gao et~al.(2015)Gao, Ver~Steeg, and Galstyan]{gao2015efficient}
Shuyang Gao, Greg Ver~Steeg, and Aram Galstyan.
\newblock Efficient estimation of mutual information for strongly dependent
  variables.
\newblock In \emph{Artificial intelligence and statistics}, pp.\  277--286,
  2015.

\bibitem[Gorham \& Mackey(2015)Gorham and Mackey]{gorham2015measuring}
Jackson Gorham and Lester Mackey.
\newblock Measuring sample quality with stein's method.
\newblock In \emph{Advances in Neural Information Processing Systems}, pp.\
  226--234, 2015.

\bibitem[Hjelm et~al.(2019)Hjelm, Fedorov, Lavoie{-}Marchildon, Grewal,
  Bachman, Trischler, and Bengio]{DBLP:conf/iclr/HjelmFLGBTB19}
R.~Devon Hjelm, Alex Fedorov, Samuel Lavoie{-}Marchildon, Karan Grewal, Philip
  Bachman, Adam Trischler, and Yoshua Bengio.
\newblock Learning deep representations by mutual information estimation and
  maximization.
\newblock In \emph{ICLR}, 2019.

\bibitem[Hu et~al.(2017)Hu, Miyato, Tokui, Matsumoto, and
  Sugiyama]{hu2017learning}
Weihua Hu, Takeru Miyato, Seiya Tokui, Eiichi Matsumoto, and Masashi Sugiyama.
\newblock Learning discrete representations via information maximizing
  self-augmented training.
\newblock In \emph{ICML}, 2017.

\bibitem[Kandasamy et~al.(2015)Kandasamy, Krishnamurthy, Poczos, Wasserman,
  et~al.]{kandasamy2015nonparametric}
Kirthevasan Kandasamy, Akshay Krishnamurthy, Barnabas Poczos, Larry Wasserman,
  et~al.
\newblock Nonparametric von mises estimators for entropies, divergences and
  mutual informations.
\newblock In \emph{Advances in Neural Information Processing Systems}, pp.\
  397--405, 2015.

\bibitem[Kinney \& Atwal(2014)Kinney and Atwal]{kinney2014equitability}
Justin~B Kinney and Gurinder~S Atwal.
\newblock Equitability, mutual information, and the maximal information
  coefficient.
\newblock \emph{Proceedings of the National Academy of Sciences}, 111\penalty0
  (9):\penalty0 3354--3359, 2014.

\bibitem[Kolchinsky et~al.(2017)Kolchinsky, Tracey, and
  Wolpert]{kolchinsky2017nonlinear}
Artemy Kolchinsky, Brendan~D Tracey, and David~H Wolpert.
\newblock Nonlinear information bottleneck.
\newblock \emph{arXiv preprint arXiv:1705.02436}, 2017.

\bibitem[Kozachenko \& Leonenko(1987)Kozachenko and
  Leonenko]{kozachenko1987sample}
LF~Kozachenko and Nikolai~N Leonenko.
\newblock Sample estimate of the entropy of a random vector.
\newblock \emph{Problemy Peredachi Informatsii}, 23\penalty0 (2):\penalty0
  9--16, 1987.

\bibitem[Kraskov et~al.(2004)Kraskov, St{\"o}gbauer, and
  Grassberger]{kraskov2004estimating}
Alexander Kraskov, Harald St{\"o}gbauer, and Peter Grassberger.
\newblock Estimating mutual information.
\newblock \emph{Physical review E}, 69\penalty0 (6):\penalty0 066138, 2004.

\bibitem[Krause et~al.(2010)Krause, Perona, and
  Gomes]{krause2010discriminative}
Andreas Krause, Pietro Perona, and Ryan~G Gomes.
\newblock Discriminative clustering by regularized information maximization.
\newblock In \emph{Advances in neural information processing systems}, 2010.

\bibitem[Krishnaswamy et~al.(2014)Krishnaswamy, Spitzer, Mingueneau, Bendall,
  Litvin, Stone, Pe’er, and Nolan]{krishnaswamy2014conditional}
Smita Krishnaswamy, Matthew~H Spitzer, Michael Mingueneau, Sean~C Bendall, Oren
  Litvin, Erica Stone, Dana Pe’er, and Garry~P Nolan.
\newblock Conditional density-based analysis of t cell signaling in single-cell
  data.
\newblock \emph{Science}, 346\penalty0 (6213):\penalty0 1250689, 2014.

\bibitem[Kwak \& Choi(2002)Kwak and Choi]{kwak2002input}
Nojun Kwak and Chong-Ho Choi.
\newblock Input feature selection by mutual information based on parzen window.
\newblock \emph{IEEE Transactions on Pattern Analysis \& Machine Intelligence},
  \penalty0 (12):\penalty0 1667--1671, 2002.

\bibitem[Li \& Turner(2017)Li and Turner]{li2017gradient}
Yingzhen Li and Richard~E Turner.
\newblock Gradient estimators for implicit models.
\newblock \emph{arXiv preprint arXiv:1705.07107}, 2017.

\bibitem[Linsker(1988)]{linsker1988self}
Ralph Linsker.
\newblock Self-organization in a perceptual network.
\newblock \emph{Computer}, 21\penalty0 (3):\penalty0 105--117, 1988.

\bibitem[Liu \& Wang(2016)Liu and Wang]{liu2016stein}
Qiang Liu and Dilin Wang.
\newblock Stein variational gradient descent: A general purpose bayesian
  inference algorithm.
\newblock In \emph{Advances in neural information processing systems}, pp.\
  2378--2386, 2016.

\bibitem[Maes et~al.(1997)Maes, Collignon, Vandermeulen, Marchal, and
  Suetens]{maes1997multimodality}
Frederik Maes, Andre Collignon, Dirk Vandermeulen, Guy Marchal, and Paul
  Suetens.
\newblock Multimodality image registration by maximization of mutual
  information.
\newblock \emph{IEEE transactions on Medical Imaging}, 16\penalty0
  (2):\penalty0 187--198, 1997.

\bibitem[McAllester \& Statos(2018)McAllester and Statos]{mcallester2018formal}
David McAllester and Karl Statos.
\newblock Formal limitations on the measurement of mutual information.
\newblock \emph{arXiv preprint arXiv:1811.04251}, 2018.

\bibitem[Moon et~al.(1995)Moon, Rajagopalan, and Lall]{moon1995estimation}
Young-Il Moon, Balaji Rajagopalan, and Upmanu Lall.
\newblock Estimation of mutual information using kernel density estimators.
\newblock \emph{Physical Review E}, 52\penalty0 (3):\penalty0 2318, 1995.

\bibitem[M{\"u}ller et~al.(2012)M{\"u}ller, Nowozin, and
  Lampert]{muller2012information}
Andreas~C M{\"u}ller, Sebastian Nowozin, and Christoph~H Lampert.
\newblock Information theoretic clustering using minimum spanning trees.
\newblock In \emph{Joint DAGM (German Association for Pattern Recognition) and
  OAGM Symposium}, pp.\  205--215. Springer, 2012.

\bibitem[Oord et~al.(2018)Oord, Li, and Vinyals]{oord2018representation}
Aaron van~den Oord, Yazhe Li, and Oriol Vinyals.
\newblock Representation learning with contrastive predictive coding.
\newblock \emph{NIPS}, 2018.

\bibitem[Palmer et~al.(2015)Palmer, Marre, Berry, and
  Bialek]{palmer2015predictive}
Stephanie~E Palmer, Olivier Marre, Michael~J Berry, and William Bialek.
\newblock Predictive information in a sensory population.
\newblock \emph{Proceedings of the National Academy of Sciences}, 112\penalty0
  (22):\penalty0 6908--6913, 2015.

\bibitem[Peng et~al.(2005)Peng, Long, and Ding]{peng2005feature}
Hanchuan Peng, Fuhui Long, and Chris Ding.
\newblock Feature selection based on mutual information: criteria of
  max-dependency, max-relevance, and min-redundancy.
\newblock \emph{IEEE Transactions on Pattern Analysis \& Machine Intelligence},
  \penalty0 (8):\penalty0 1226--1238, 2005.

\bibitem[Peng et~al.(2019)Peng, Kanazawa, Toyer, Abbeel, and
  Levine]{DBLP:conf/iclr/PengKTAL19}
Xue~Bin Peng, Angjoo Kanazawa, Sam Toyer, Pieter Abbeel, and Sergey Levine.
\newblock Variational discriminator bottleneck: Improving imitation learning,
  inverse rl, and gans by constraining information flow.
\newblock In \emph{7th International Conference on Learning Representations,
  {ICLR} 2019, New Orleans, LA, USA, May 6-9, 2019}, 2019.

\bibitem[P{\'e}rez-Cruz(2008)]{perez2008kullback}
Fernando P{\'e}rez-Cruz.
\newblock Kullback-leibler divergence estimation of continuous distributions.
\newblock In \emph{2008 IEEE international symposium on information theory},
  pp.\  1666--1670. IEEE, 2008.

\bibitem[Poole et~al.(2019)Poole, Ozair, van~den Oord, Alemi, and
  Tucker]{DBLP:conf/icml/PooleOOAT19}
Ben Poole, Sherjil Ozair, A{\"{a}}ron van~den Oord, Alex Alemi, and George
  Tucker.
\newblock On variational bounds of mutual information.
\newblock In \emph{Proceedings of the 36th International Conference on Machine
  Learning, {ICML} 2019, 9-15 June 2019, Long Beach, California, {USA}}, pp.\
  5171--5180, 2019.

\bibitem[Roeder et~al.(2017)Roeder, Wu, and Duvenaud]{roeder2017sticking}
Geoffrey Roeder, Yuhuai Wu, and David~K Duvenaud.
\newblock Sticking the landing: Simple, lower-variance gradient estimators for
  variational inference.
\newblock In \emph{Advances in Neural Information Processing Systems}, pp.\
  6925--6934, 2017.

\bibitem[Shi et~al.(2018)Shi, Sun, and Zhu]{shi2018spectral}
Jiaxin Shi, Shengyang Sun, and Jun Zhu.
\newblock A spectral approach to gradient estimation for implicit
  distributions.
\newblock \emph{arXiv preprint arXiv:1806.02925}, 2018.

\bibitem[Shwartz-Ziv \& Tishby(2017)Shwartz-Ziv and Tishby]{shwartz2017opening}
Ravid Shwartz-Ziv and Naftali Tishby.
\newblock Opening the black box of deep neural networks via information.
\newblock \emph{arXiv preprint arXiv:1703.00810}, 2017.

\bibitem[Singh \& P{\'o}czos(2016)Singh and P{\'o}czos]{singh2016finite}
Shashank Singh and Barnab{\'a}s P{\'o}czos.
\newblock Finite-sample analysis of fixed-k nearest neighbor density functional
  estimators.
\newblock In \emph{Advances in neural information processing systems}, pp.\
  1217--1225, 2016.

\bibitem[Slonim \& Tishby(2000)Slonim and Tishby]{slonim2000document}
Noam Slonim and Naftali Tishby.
\newblock Document clustering using word clusters via the information
  bottleneck method.
\newblock In \emph{Proceedings of the 23rd annual international ACM SIGIR
  conference on Research and development in information retrieval}, pp.\
  208--215. ACM, 2000.

\bibitem[Song et~al.(2019)Song, Garg, Shi, and Ermon]{song2019sliced}
Yang Song, Sahaj Garg, Jiaxin Shi, and Stefano Ermon.
\newblock Sliced score matching: A scalable approach to density and score
  estimation.
\newblock \emph{arXiv preprint arXiv:1905.07088}, 2019.

\bibitem[Suzuki et~al.(2008)Suzuki, Sugiyama, Sese, and
  Kanamori]{suzuki2008approximating}
Taiji Suzuki, Masashi Sugiyama, Jun Sese, and Takafumi Kanamori.
\newblock Approximating mutual information by maximum likelihood density ratio
  estimation.
\newblock In \emph{New challenges for feature selection in data mining and
  knowledge discovery}, pp.\  5--20, 2008.

\bibitem[Tishby \& Zaslavsky(2015)Tishby and Zaslavsky]{tishby2015deep}
Naftali Tishby and Noga Zaslavsky.
\newblock Deep learning and the information bottleneck principle.
\newblock In \emph{2015 IEEE Information Theory Workshop (ITW)}, pp.\  1--5.
  IEEE, 2015.

\bibitem[Tishby et~al.(2000)Tishby, Pereira, and Bialek]{tishby2000information}
Naftali Tishby, Fernando~C Pereira, and William Bialek.
\newblock The information bottleneck method.
\newblock \emph{arXiv preprint physics/0004057}, 2000.

\bibitem[Tschannen et~al.(2019)Tschannen, Djolonga, Rubenstein, Gelly, and
  Lucic]{Tschannen2019OnMI}
Michael Tschannen, Josip Djolonga, Paul~K. Rubenstein, Sylvain Gelly, and Mario
  Lucic.
\newblock On mutual information maximization for representation learning.
\newblock \emph{ArXiv}, abs/1907.13625, 2019.

\bibitem[Ver~Steeg \& Galstyan(2015)Ver~Steeg and Galstyan]{ver2015maximally}
Greg Ver~Steeg and Aram Galstyan.
\newblock Maximally informative hierarchical representations of
  high-dimensional data.
\newblock In \emph{Artificial Intelligence and Statistics}, pp.\  1004--1012,
  2015.

\bibitem[Vera et~al.(2018)Vera, Piantanida, and Vega]{vera2018role}
Matias Vera, Pablo Piantanida, and Leonardo~Rey Vega.
\newblock The role of the information bottleneck in representation learning.
\newblock In \emph{IEEE International Symposium on Information Theory (ISIT)},
  pp.\  1580--1584. IEEE, 2018.

\bibitem[Xu et~al.(2015)Xu, Jin, Shen, and Zhu]{DBLP:conf/aaai/XuJSZ15}
Zenglin Xu, Rong Jin, Bin Shen, and Shenghuo Zhu.
\newblock Nystrom approximation for sparse kernel methods: Theoretical analysis
  and empirical evaluation.
\newblock In Blai Bonet and Sven Koenig (eds.), \emph{Proceedings of the
  Twenty-Ninth {AAAI} Conference on Artificial Intelligence, January 25-30,
  2015, Austin, Texas, {USA}}, pp.\  3115--3121. {AAAI} Press, 2015.
\newblock URL
  \url{http://www.aaai.org/ocs/index.php/AAAI/AAAI15/paper/view/9860}.

\bibitem[Zeitler et~al.(2008)Zeitler, Koetter, Bauch, and
  Widmer]{zeitler2008design}
Georg Zeitler, Ralf Koetter, Gerhard Bauch, and Joerg Widmer.
\newblock Design of network coding functions in multihop relay networks.
\newblock In \emph{2008 5th International Symposium on Turbo Codes and Related
  Topics}, pp.\  249--254. IEEE, 2008.

\end{thebibliography}
\bibliographystyle{iclr2020_conference}

\newpage
\appendix

\section{Derivation of Gradient Estimates for  Entropy}
\label{A_entropy}
\textbf{Unconditional Entropy}
Given that the encoder $E_{\psi}(.)$ is deterministic, our goal is to optimize the entropy $\mathrm{H}(q)=-\mathbb{E}_{q} \log q$, where $q$ is short for the distribution $q_{\psi}(\mathbf{z})$ of the representation $\mathbf{z}$ w.r.t. its parameters $\psi$. We can decompose the gradient of the entropy of $q_{\psi}(\mathbf{z})$ as:
\begin{align}\label{gentropy}
 \nabla_{\psi} H(z) &= - \nabla_{\psi} \mathbb{E}_{ q_{\psi}(\mathbf{z})} [\log q(\mathbf{z})]- \mathbb{E}_{ q(\mathbf{z})} [\nabla_{\psi}\log q_{\psi}(\mathbf{z})],
\end{align}

The second term on the right side of the equation can be calculated:
\begin{align}
 \mathbb{E}_{ q(\mathbf{z})} [\nabla_{\psi}\log q_{\psi}(\mathbf{z})]
 = \mathbb{E}_{ q(\mathbf{z})} [\nabla_{\psi}q_{\psi}(\mathbf{z})\times \frac{1}{q(\mathbf{z})}]
 = \int \nabla_{\psi}q_{\psi}(\mathbf{z})d \mathbf{z}
 = \nabla_{\psi}\int q_{\psi}(\mathbf{z})d \mathbf{z}
 =0.
\end{align}
Therefore, the gradient of the entropy of $q_{\psi}(\mathbf{z})$ becomes
\begin{align}
 \nabla_{\psi} H(z) &=  -\nabla_{\psi} \mathbb{E}_{ q_{\psi}(\mathbf{z})} [\log q(\mathbf{z})].
\end{align}

\textbf{Conditional Entropy}
Consider nondeterministic encoder function $E_{\psi}(.,\boldsymbol{\epsilon})$ where $\boldsymbol{\epsilon}$ is an auxiliary variable with independent marginal $p(\boldsymbol{\epsilon})$. The distribution $q_{\psi}(z|x)$ is determined by $\boldsymbol{\epsilon}$ and the encoder parameters $\psi$. The auxiliary variable $\boldsymbol{\epsilon}$ introduces randomness to the encoder. 
First, we decompose the gradients of Conditional Entropy as following:
\begin{align}\label{conetrp}
\nabla_{\psi}H\left(\mathbf{z|x}\right) &=-\nabla_{\psi}\int p_{\psi}(\mathbf{z,x}) \log p_{\psi}(\mathbf{z|x}) d z d x \nonumber\\
&=-\mathbb{E}_{p(\mathbf{x})}[\nabla_{\psi}\int p_{\psi}(\mathbf{z|x}) \log p_{\psi}(\mathbf{z|x}) d z] \nonumber\\
&=-\mathbb{E}_{p(\mathbf{x})}[\nabla_{\psi} \mathbb{E}_{p_{\psi}(\mathbf{z|x})} [\log p(\mathbf{z|x})]+\int p(\mathbf{z|x}) \nabla_{\psi}\log p_{\psi}(\mathbf{z|x}) d h] \nonumber\\
&=-\mathbb{E}_{p(\mathbf{x})}[\nabla_{\psi} \mathbb{E}_{p_{\psi}(\mathbf{z|x})} [\log p(\mathbf{z|x})]+\int \nabla_{\psi} p_{\psi}(\mathbf{z|x})d h]\nonumber\\
&=-\mathbb{E}_{p(\mathbf{x})}[\nabla_{\psi} \mathbb{E}_{p_{\psi}(\mathbf{z|x})} [\log p(\mathbf{z|x})]- \nabla_{\psi}\int p_{\psi}(\mathbf{h,x})d h d x]\nonumber\\
&=-\mathbb{E}_{p(\mathbf{x})}[\nabla_{\psi} \mathbb{E}_{p_{\psi}(\mathbf{z|x})} [\log p(\mathbf{z|x})]].
\end{align}
Note that $\mathbf{z}=E_\psi(\mathbf{x},\boldsymbol{\epsilon})$,  such that  we can apply reparameterization trick to the gradient estimator of conditional entropy in Eq. (\ref{conetrp}),
\begin{align}
  H_{\psi}(\mathbf{z}|\mathbf{x})=-\mathbb{E}_{p(\mathbf{x})}[\mathbb{E}_{p(\boldsymbol{\epsilon})}[\nabla_{(\mathbf{z}|\mathbf{x})}\log q(E_\psi(\mathbf{x},\boldsymbol{\epsilon})|\mathbf{x})\nabla_{\psi}E_\psi(\mathbf{x},\boldsymbol{\epsilon})]].
\end{align}
\section{MIGE Algorithm Description}
    The algorithm description of our proposed MIGE is stated in Algorithm \ref{alg}.
\begin{algorithm}[H]
\caption{MIGE (Circumstance I)}
\label{alg}
\begin{algorithmic}
    \STATE{\textbf{1. Sampling}:\\
     ~~~~~~ Draw $n$ samples from the data distribution $p(x)$, $n$ denotes mini-batch size, \\
         ~~~~~~ then compute the corresponding output of the encoder \\~~~~~~
    $(\mathbf{x}^{(1)},\mathbf{z}^{(1)}),\cdots,(\mathbf{x}^{(n)},\mathbf{z}^{(n)}) \sim q_\psi (\mathbf{x},\mathbf{z})$\\~~~~~~
    $\mathbf{z}^{(1)},\cdots,\mathbf{z}^{(n)}\sim q_\psi (\mathbf{z})$}
    \STATE{\textbf{2. Estimate the score function}:\\~~~~~~
    $\nabla_{\mathbf{z}} \log q_\psi (\mathbf{z}^{(i)}) \leftarrow \mathrm{SSGE}(\mathbf{z}^{(1)},\cdots,\mathbf{z}^{(n)})$\\~~~~~~
    $\nabla_{(\mathbf{x},\mathbf{z})} \log q_\psi (\mathbf{x}^{(i)},\mathbf{z}^{(i)}) \leftarrow \mathrm{SSGE}((\mathbf{x}^{(1)},\mathbf{z}^{(1)}),\cdots,(\mathbf{x}^{(n)},\mathbf{z}^{(n)}))$
    }
    \STATE{\textbf{3. Estimate the entropy gradient}:\\~~~~~~
    $\nabla_\psi H(\mathbf{z}) \leftarrow - \frac{1}{n}  \sum_{i=1}^{n} \left[ \nabla_\psi \mathbf{z}^{(i)} \nabla_{\mathbf{z}} \log q_\psi (\mathbf{z}^{(i)}) \right]$ \\~~~~~~
    $\nabla_\psi H(\mathbf{x}, \mathbf{z}) \leftarrow - \frac{1}{n}  \sum_{i=1}^{n} \left[\nabla_\psi (\mathbf{x}^{(i)},\mathbf{z}^{(i)}) \nabla_{(\mathbf{x},\mathbf{z})} \log q_\psi (\mathbf{x}^{(i)},\mathbf{z}^{(i)}) \right]$
    }
    \STATE{\textbf{4. Estimate the MI gradient}:\\~~~~~~
    $\nabla_\psi I(\mathbf{x};\mathbf{z}) \leftarrow \nabla_\psi H(\mathbf{z}) - \nabla_\psi H(\mathbf{x};\mathbf{z})$
    }
 
\end{algorithmic}
\end{algorithm}

\section{Discussion on DIM(L)}
DIM(L) \citep{DBLP:conf/iclr/HjelmFLGBTB19} is the state-of-the-art unsupervised model for representaion learning, which maximizes the average MI between the high-level representation and local patches of the image, and achieve an even higher classification accuracy than supervised learning. As shown in Table \ref{table:infomax_diml}, we apply MIGE into DIM(L) and surprisingly find there is a significant performance gap to DIM(L). 

To our knowledge, the principle of DIM(L) is still unclear. \cite{Tschannen2019OnMI} argues that maximizing tighter bounds in DIM(L) can lead to worse results, and the success of these methods cannot be attributed to the properties of MI alone, and they strongly depend on the inductive bias in both the choice of feature extractor architectures and the parameterization of the employed MI estimators. For MIGE, we are investigating the behind reasons, e.g., to investigate the distributions of the patches.

\begin{table}[H]
\caption{CIFAR-10 and CIFAR-100 classification accuracy (top 1) of downstream tasks compared with vanilla DIM(L).}
    \label{table:infomax_diml}
    \begin{center}
    \begin{tabular}{c||ccc|ccc}
    \multirow{2}{*}{\textbf{Model}} & \multicolumn{3}{c|}{\textbf{CIFAR-10}} & \multicolumn{3}{c}{\textbf{CIFAR-100}} \\ 
                        & conv           & fc(1024)       & Y(64)          & conv           & fc(1024)       & Y(64) \\ \hline 
    % Fully supervised    & \multicolumn{3}{c|}{75.39}                      & \multicolumn{3}{c}{42.27}               \\ \hline
    DIM(L) (JSD)                 & 72.16\%        & 67.99\%       & 66.35\%          & 41.65\%          & 39.60\%          & 39.66\% \\
    DIM(L)  (JSD + PM)            & 73.25\%           & 73.62\%          & 66.96\%          & 48.13\%          & 45.92\%          & 39.6\% \\
    DIM(L)  (infoNCE)             & 75.05\%          & 70.68\%          & \textbf{69.24}\%          & 44.11\%          & 42.97\%          & \textbf{42.74}\% \\
    DIM(L)  (infoNCE + PM)        & \textbf{75.21}\%          & \textbf{75.57\%}         & 69.13\%          & \textbf{49.74}\%          & \textbf{47.72}\%          & 41.61\% \\ \hline
    MIGE             & 59.72\% & 56.14\% & 54.01\% & 30.0\% & 28.96\% & 27.65\% \\ \hline
    \end{tabular}
    \end{center}
\end{table}

\end{document}